\documentclass{article}

\usepackage{microtype}
\usepackage{graphicx}
\usepackage{subfigure}
\usepackage{amssymb}
\usepackage{booktabs} 

\usepackage{soul}

\usepackage{glossaries}
\newacronym{rl}{RL}{Reinforcement learning}

\makeatletter
\g@addto@macro\@floatboxreset{\centering}
\makeatother

\usepackage{hyperref}

\usepackage[accepted]{icml2020}

\icmltitlerunning{Pre-trained Word Embeddings for Goal-conditional Transfer Learning in Reinforcement Learning}

\begin{document}

\twocolumn[

\icmltitle{Pre-trained Word Embeddings for Goal-conditional Transfer Learning in Reinforcement Learning}




\begin{icmlauthorlist}
\icmlauthor{Matthias Hutsebaut-Buysse}{ua}
\icmlauthor{Kevin Mets}{ua}
\icmlauthor{Steven Latr\'e}{ua}
\end{icmlauthorlist}

\icmlaffiliation{ua}{Department of Computer Science, University of Antwerp - imec}

\icmlcorrespondingauthor{Matthias Hutsebaut-Buysse}{matthias.hutsebaut-buysse@uantwerpen.be}

\icmlkeywords{Machine Learning, ICML}

\vskip 0.3in
]




\printAffiliationsAndNotice

\begin{abstract}
\gls{rl} algorithms typically start \textit{tabula rasa}, without any prior knowledge of the environment, and without any prior skills. This however often leads to low sample efficiency, requiring a large amount of interaction with the environment. This is especially true in a lifelong learning setting, in which the agent needs to continually extend its capabilities. In this paper, we examine how a pre-trained task-independent word embedding can make a goal-conditional \gls{rl} agent more sample efficient. We do this by facilitating transfer learning between different related tasks. We experimentally demonstrate our approach on a set of object navigation tasks.
\end{abstract}

\section{Introduction}

In order to build complex intelligent systems, an agent needs to be capable of re-using and adapting previously learned traits. This property is often called the \textit{learning-to-learn} \cite{lake2017building_ai} ability of an agent.

This \textit{learning-to-learn} approach is however in sharp contrast to how most \gls{rl} approaches \citep{openai2020agent57, kapturowski2019r2d2, schrittwieser2019muzero} currently are capable of solving sequential decision-making problems. Current \gls{rl} algorithms typically start \textit{tabula rasa}, and do not re-use any knowledge previously learned in past tasks. These approaches are often very sample inefficient, requiring an unreasonable amount of interaction with the environment in order to learn new tasks.

A \textit{learning-to-learn} approach could allow the agent to become more sample efficient, by allowing the agent to build upon what it already learned in past similar tasks. However, how to implement \textit{learning-to-learn} in \gls{rl} has remained mostly an open question. In supervised machine learning with neural networks, training performance on vision tasks can be significantly increased by re-using the initial layers of a previously trained neural network. These initial layers learn to recognize features which are mostly task-independent \citep{yosinski2014transfer_features}. Layers on top of these features learn to map combinations of the resulting features to the output labels. 

Similar approaches have been used in \gls{rl} \citep{taylor2009transfer_rl_survey}. Especially in \textit{deep} \gls{rl}, when working with high-dimensional inputs, it makes a lot of sense to re-use parts of the (learned) visual pipeline across different tasks \citep{chaplot2016TransferDeepReinforcement}, especially when this high-dimensional input is visually complex.

However, mapping a high-dimensional input to a latent representation, is only part of the \gls{rl} problem. In \gls{rl}, the agent also needs to explore the environment in order to map actions to states. Such an action can consist of performing a single primitive action, such as \textit{take one step forward}. However, exploration has been demonstrated \citep{jinnai2020deep_covering_options, eysenbach2019diayn, bacon2017oc} to be significantly faster when also utilizing temporal abstractions. These abstractions utilize multiple primitive actions, when exploring the environment (e.g.\ \textit{walk to the garden}). 

In our approach, we demonstrate that prior knowledge of a deep \gls{rl} agent can be used as temporal abstractions in order to facilitate transfer learning to a novel previously unseen tasks. We do this by utilizing a goal-conditional agent. In this style, the \gls{rl} agent receives a combination of the current state and a goal as its input. Assuming a finite set of possible goals, this goal is typically represented using a \textit{one-hot} encoded vector. In this one-hot goal-space the distance has no meaning, as the distance between different goals is always the same.

We express the goal of the agent using natural language. We do this by using a task-independent pre-trained word embedding. This allows the agent to quickly link a new, previously unseen goal to what it has already learned from past tasks. We experimentally demonstrate that these kinds of pre-trained word goal-embeddings can be used to transfer knowledge in the form of temporal abstractions in a transfer learning settings.

\section{Related work}
\Gls{rl} has been informed by natural language in various ways \citep{luketina2019suvery_rl_nlp}. The majority of research has been conducted on how language instructions can be linked to actions \citep{chenLearningInterpretNatural, mei2016ListenAttendWalk, hermann2017GroundedLanguageLearning}. Additionally, language has been used as an instrument to communicate domain knowledge \citep{zhong2019RTFM}, or assist by shaping the reward function \citep{bahdanau2019LearningUnderstandGoal}.

Similar to our work, natural language has also been used to transfer knowledge. For example in \citep{narasimhan2018language_transfer} a method is proposed to transfer knowledge between different environments. In previous work \citep{hutsebautbuysse2019fast}, we proposed a method to train a custom goal word embedding based on transfer performance.

\section{Object navigation task setting}

\begin{figure} 
    \includegraphics[scale=0.3]{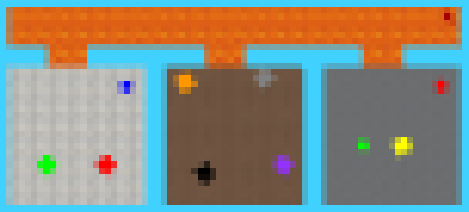}
    \caption{Top-down layout of the environment used in our experiments. The three rooms (bathroom, kitchen, bedroom) are connected through a long corridor.}
    \label{figure:env_top_down}
\end{figure}

\begin{figure} 
    \includegraphics[scale=0.15]{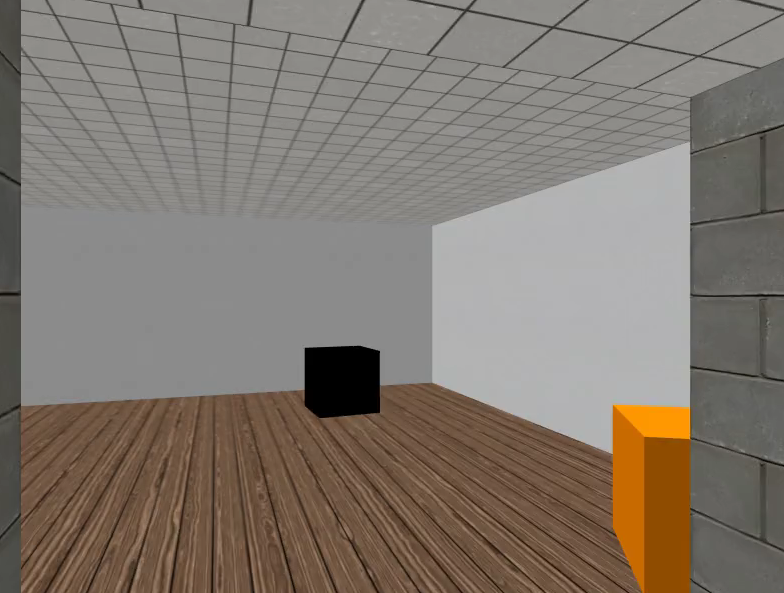}
    \caption{An example rendering of the viewpoint the agent receives as part of its state.}
    \label{figure:env_state}
\end{figure}

In this paper, we are concerned with the problem of object navigation. In a single instance of this problem, the agent is randomly spawned in a corridor, and needs to navigate towards an up-front specified object in the environment. The episode is considered successful if the agent has positioned itself near the goal object in a maximum of 500 steps.

In order to solve this problem, the agent does not have access to a map of the environment, and only needs to rely on RGB sensory input.

For our experiments, we use a custom designed level in the \textit{MiniWorld} \citep{gym_miniworld} benchmarking environment. Figure~\ref{figure:env_top_down} shows the layout of the used environment. Figure~\ref{figure:env_state} renders an example viewpoint of the agent.

Our designed level mimics a small domestic apartment. The layout  consists of three rooms connected through a corridor. Each room contains a number of objects in fixed positions:

\begin{itemize}
    \item \textbf{Bathroom}: shower, bathtub, toilet
    \item \textbf{Kitchen}: stove, toaster, table, microwave
    \item \textbf{Bedroom}: bed, wardrobe, nightstand
\end{itemize}

Objects are represented with spheres and cubes in different arbitrarily chosen colors. For example, in Figure~\ref{figure:env_state}, the black box represents the \textit{table} object.

After taking an action, the agent receives a reward which is equal to the improvement of the distance to the goal object. A \textit{slack} penalty of -0.01 is added to the reward, in order to force the agent to move. A bonus reward of 10 is awarded when reaching the goal object.

\section{Method}

\subsection{Goal-encoding}

In order for a \gls{rl} agent to be capable of executing multiple tasks, the required task can be specified to the agent using a goal-vector. In our problem setting, this goal-vector should correspond with the object the agent needs to navigate to.

Typically, in order to encode different goals, a discrete \textit{one-hot} encoding is used. Unfortunately, when using such a vector, the number of goals should be known in advance, as it is not straightforward to alter a neural network which depends on this vector.

However, in a lifelong learning setting \citep{silver2013lifelong_learning} we would like the agent to be capable of learning to navigate to new goals, without having to explicitly define the number of goals in advance. In order to support this, we propose to encode goals using a pre-trained word embedding.

Such a model is trained \citep{mikolov2013word2vec} by taking as input a large corpus of texts, and outputs a vector space. Words that appear in similar contexts, are trained to be also close to each other in the output vector space. We reason that this prior knowledge can be of great use in a multi-task object navigation task, and that goals closer in word vector space, will also transfer better between different \gls{rl} policies.

The pre-trained model we use \citep{spacy2} is trained on the OntoNotes 5 \citep{weischedelOntoNotesLargeTraining} dataset. This dataset contains a large set of different type of documents, and is not linked in any way with our task setting. The resulting model is capable of expressing a goal description with continuous vectors of size 300. 

\subsection{Training architecture}

\begin{figure}
    \includegraphics[scale=0.6]{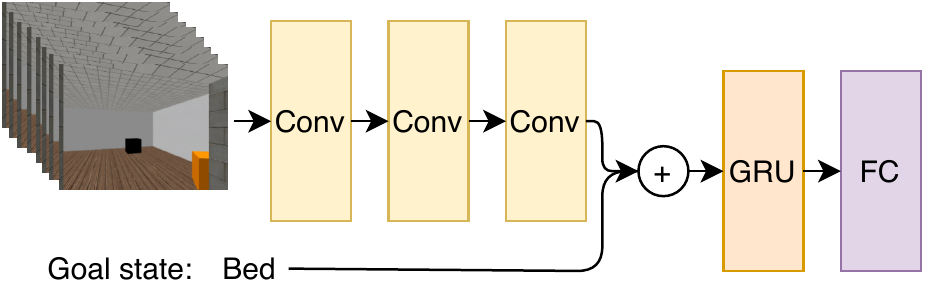}
    \caption{Goal-conditional architecture.}
    \label{img:architecture}
\end{figure}

In order to allow our agent to solve object navigation tasks, we use a standard DRQN architecture \citep{hausknecht2015rdqn}. We use the \textit{recurrent} flavor (with sequence-length 8) of the DQN algorithm \citep{mnih2015dqn}, because the current state does not contain enough information for the agent to successfully navigate the environment. The goal-vector is concatenated with the visual perception part of our architecture. This architecture is displayed in Figure~\ref{img:architecture}.

\subsection{Transfer}
\label{sec:transfer_method}
In order for our lifelong learning agent to be capable of transferring knowledge from one task to another task, we propose to adapt the $\epsilon$-greedy exploration scheme \citep{watkins1989learning}. In this scheme, the agent takes a random action $\epsilon$-percent of the time, instead of greedily following the current policy $\pi$. This allows the agent to explore (potentially better) actions, it would normally not take under the current policy. This $\epsilon$ value is typically decayed during training as the agent becomes more confident in its policy. 

We propose to instead of purely taking random exploratory actions in order to navigate to a new goal (e.g.\ \textit{bathtub}), to also explore actions which would correspond to the action the agent would take if it would be provided with a different goal-vector which the agent already has mastered before (e.g.\ \textit{shower}).

However, how can the agent know which goal-vector will transfer best to satisfy the new unseen goal? We propose to solve this question by measuring the cosine similarity of the unseen goal object and the mastered goal objects in their word embedding space. As these embeddings are trained to put words which are often related to each other close to each other in the vector-space, we reason that goals close in this space will most commonly also be located in similar positions in typical building layouts.

Intuitively using knowledge from a prior object goal allows the agent to use this knowledge as a form of temporal abstraction, which corresponds to navigating to the room the object can most likely be found.

It however remains essential that the agent keeps doing enough exploration, especially in states close to the prior goal object. We propose to introduce a sampling rate hyperparameter $\alpha$ in order to balance the trade-off between biased sampling from the prior policy, and random exploration.

In summary the policy of our agent when tasked with reaching goal $z$, word embedding $\mathcal{M}$ and prior goals $\omega_{0...i}\in\Omega$ looks as follows:
\begin{itemize}
    \item $P(1-\epsilon)$: take greedy action $\pi(s_t, z)$
    \item $P(\epsilon*\alpha)$: sample action from $\pi(s,\omega)$ with $w = argmax_w (cos(M(z), M(w)) $
    \item $P(\epsilon*(1-\alpha))$: take random action
\end{itemize}

\section{Experiments and results}

Experiments are terminated after reaching a success rate of 0.95 on the last 100.000 steps (and only minimal exploration $\epsilon=0.01$ is done ). In all experiments $\epsilon$ is linearly decayed over 1M steps, and we use an experience replay buffer of size 500.000.

\subsection{Using language goal-vector vs one-hot goal-vector}
In our first experiment, we examine the impact of the goal-vector on the training performance when training a goal-conditional agent on a set of four different goals.

\begin{figure}
    \includegraphics[scale=0.48]{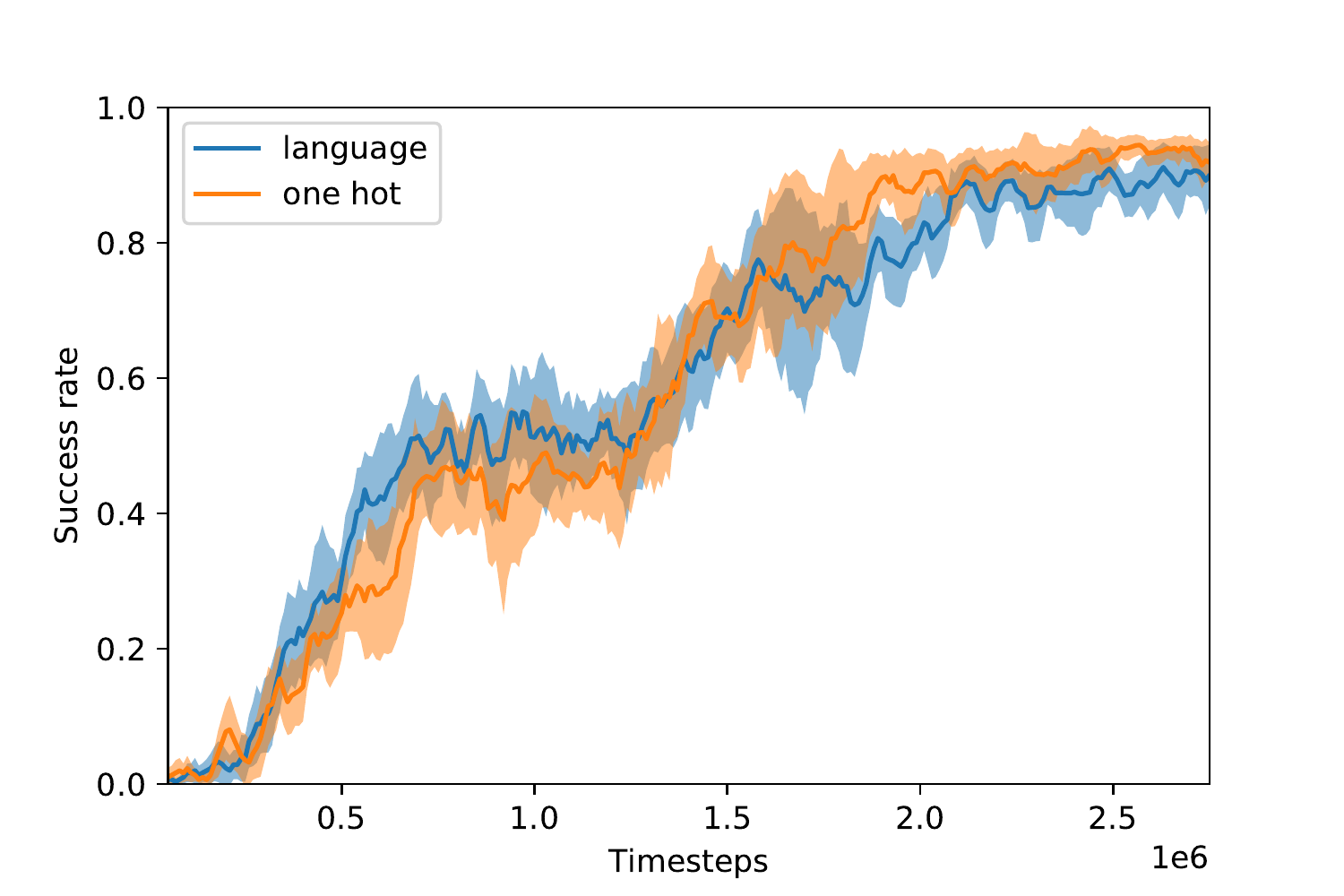}
    \caption{Comparing one-hot encoding vs language goal-vector on a set of 4 goals (results are averaged over 3 runs).}
    \label{plot:lang_vs_1h}
\end{figure}

The results of our experiments presented in Figure~\ref{plot:lang_vs_1h} give an indication that directly specifying the goal object using the word embedding ($\mathbb{R}^{300}$) has no significant negative effect over using a one-hot goal object encoding ($\mathbb{R}^{10}$). There also seems to be an interesting relation that using the word goal descriptions has a slightly positive effect of exploration, and using the one-hot encoding seems to work better when the policy is almost ($\epsilon=0.01$) completely greedy (after 1M timesteps).

Using the goal word embedding for our lifelong learning agent is ideal, as we do not need to specify the amount of possible goal objects up front. The word goal embedding allows us to input a large amount of goals (the used model has 20k unique vectors).

\subsection{Initial training on limited goal sets}

\begin{figure}[ht]
    \includegraphics[scale=0.48]{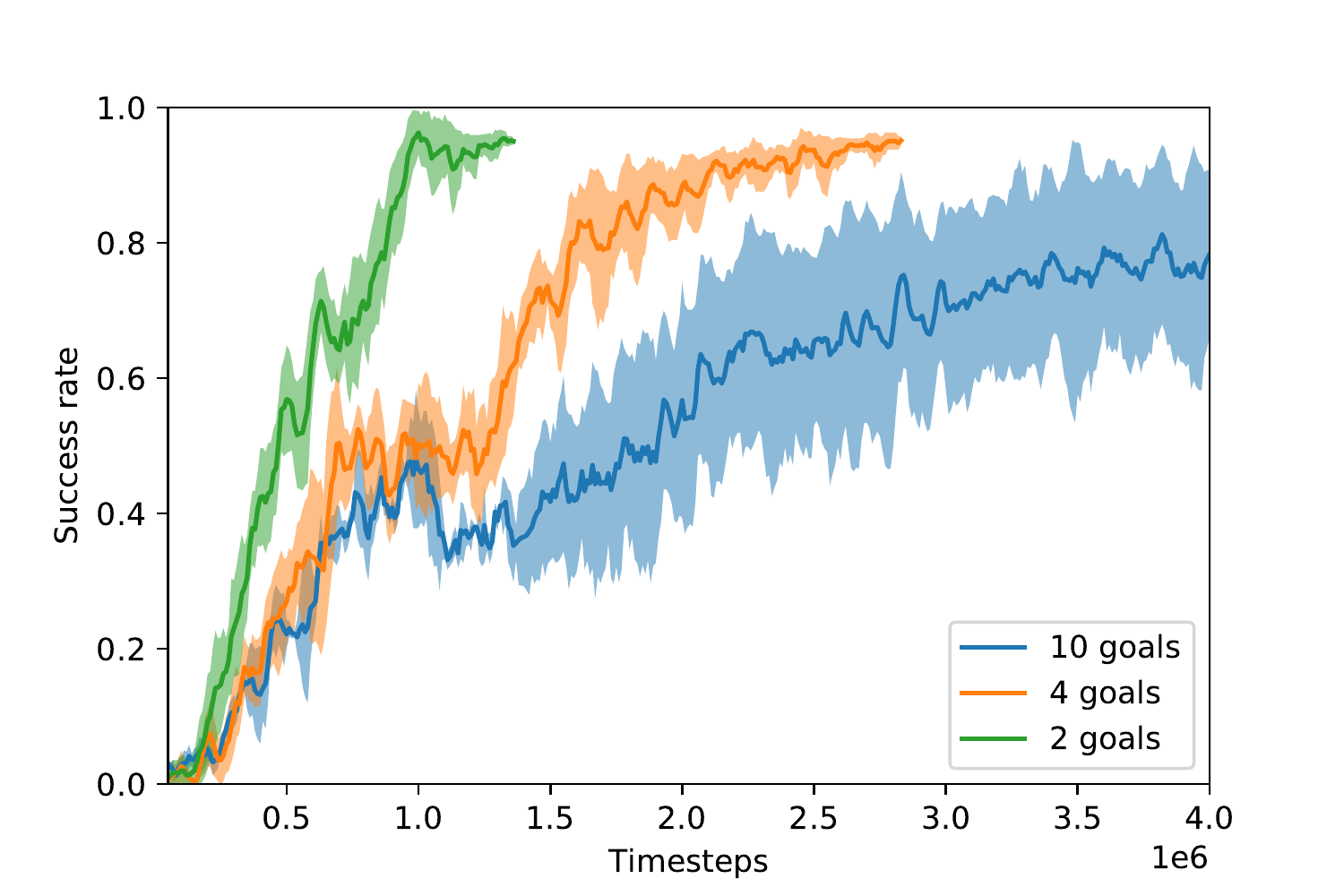}
    \caption{Comparing training performance on different sizes of goal object sets (results are averaged over 3 runs).}
    \label{plot:base_performance}
\end{figure}

We would like our lifelong agent to be capable of navigating to as many goals as possible. In order to do so, we could train our agent on a large set of goals. However, research has demonstrated \citep{narvekar2020curriculum_learning} that using a carefully selected task curriculum often leads to better results.

We plotted the results of training our agent using different sizes of goal sets, in Figure~\ref{plot:base_performance}. These results demonstrate that larger sets of goals are significantly harder to train. This finding supports our claim that a lifelong learning agent significantly benefits from first learning a small sub-set of goals, and gradually expanding its capabilities through transfer learning.

\subsection{Transfer to new objects using prior policy}
In our final experiment, we allow the agent to transfer knowledge from one goal object to a different unseen goal object using the transfer mechanism described in Section~\ref{sec:transfer_method}.

We start with a policy which has been trained to reliably reach four goal objects in the environment (\textit{shower, toilet, bed} and \textit{toaster}). In this experiment we test the transfer capability of our algorithm in order to learn to reach a new goal object \textit{bathtub} using a prior sampling-rate of $\alpha=0.2$. The new policy is randomly initialized.

\begin{figure}[ht]
    \includegraphics[scale=0.42]{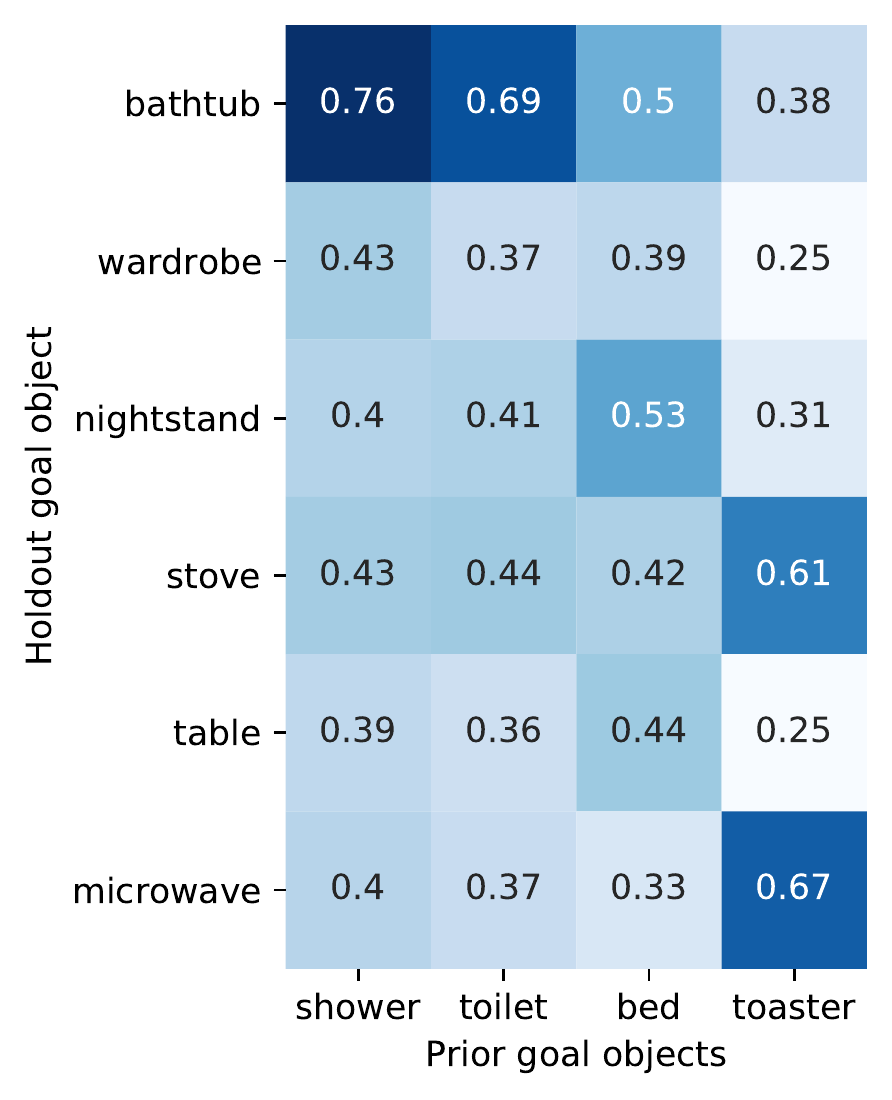}
    \caption{Cosine similarity of holdout goal objects and prior goal objects in the word embedding}
    \label{plot:goal_similarity}
\end{figure}

\begin{figure}
    \includegraphics[scale=0.48]{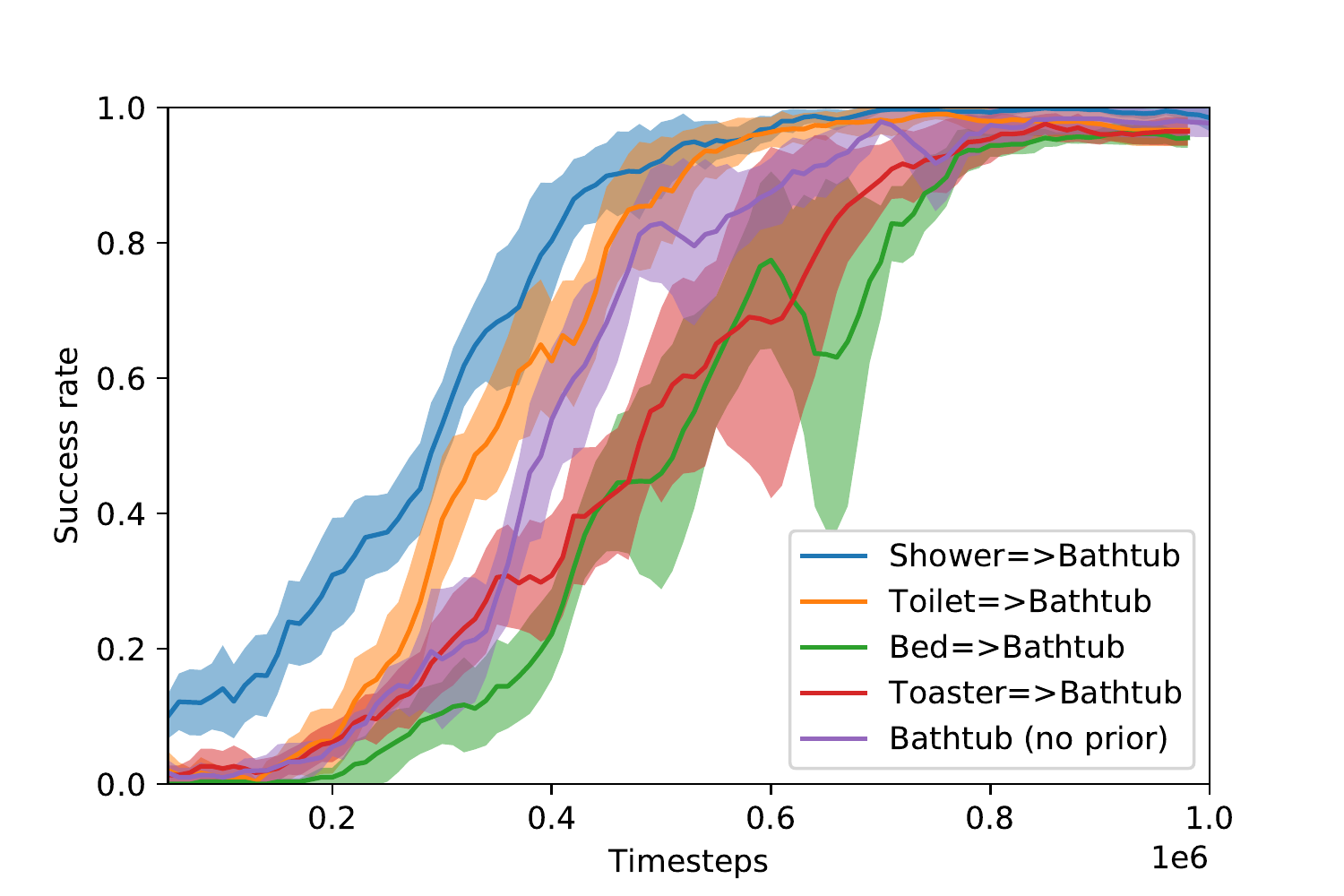}
    \caption{Comparison of using different prior goals in order to learn how to reach a new unseen goal object (\textit{bathtub}) (results are averaged over 3 runs).}
    \label{plot:transfer_example}
\end{figure}

Our preliminary results, plotted in Figure~\ref{plot:transfer_example}, demonstrate that the goal object that transfers best to the new unseen goal object (\textit{bathtub}) is \textit{shower}, which is also the goal object that is closest in language space (Figure~\ref{plot:goal_similarity}). The performance when using the second closest goal ($\textit{toilet}$) in language space also performs similarly. 

Unrelated goals such as \textit{bed} and \textit{toaster} hinder the agent, steering the agent to the wrong room (\textit{kitchen}) and we observe a negative transfer effect compared to just learning to navigate to the goal without any prior knowledge.

\section{Conclusion}
In this paper we presented our preliminary ideas on how natural language can assist a \gls{rl} agent in a lifelong learning setting.

Our approach consists of training the agent on small sets of goals, directly inputting the goal descriptions in natural language. We utilize similarity of descriptions of seen and unseen goal objects in natural language in order to decide how to transfer existing knowledge to novel tasks. In order to transfer knowledge, we propose a simple, but effective transfer mechanism.

We support our method with preliminary results in a 3D simulated domestic environment. In future work we propose to further examine the impact of different language models, utilize more complex floor layouts, and we would like to study more complex prior goal selection schemes (e.g.\ use different prior goals weighted by their similarity with the current goal).

\section*{Acknowledgements}
This research received funding from the Flemish Government under the ``Onderzoeksprogramma Artificiële Intelligentie (AI) Vlaanderen'' programme.

\bibliography{pretrained_lang}

\begin{thebibliography}{27}
\providecommand{\natexlab}[1]{#1}
\providecommand{\url}[1]{\texttt{#1}}
\expandafter\ifx\csname urlstyle\endcsname\relax
  \providecommand{\doi}[1]{doi: #1}\else
  \providecommand{\doi}{doi: \begingroup \urlstyle{rm}\Url}\fi

\bibitem[Bacon et~al.(2017)Bacon, Harb, and Precup]{bacon2017oc}
Bacon, P.-L., Harb, J., and Precup, D.
\newblock The {{Option}}-{{Critic Architecture}}.
\newblock In \emph{{{AAAI17}}}, 2017.

\bibitem[Badia et~al.(2020)Badia, Piot, Kapturowski, Sprechmann, Vitvitskyi,
  Guo, and Blundell]{openai2020agent57}
Badia, A.~P., Piot, B., Kapturowski, S., Sprechmann, P., Vitvitskyi, A., Guo,
  D., and Blundell, C.
\newblock Agent57: {{Outperforming}} the {{Atari Human Benchmark}}.
\newblock 2020.

\bibitem[Bahdanau et~al.(2019)Bahdanau, Hill, Leike, Hughes, Hosseini, Kohli,
  and Grefenstette]{bahdanau2019LearningUnderstandGoal}
Bahdanau, D., Hill, F., Leike, J., Hughes, E., Hosseini, A., Kohli, P., and
  Grefenstette, E.
\newblock Learning to {{Understand Goal Specifications}} by {{Modelling
  Reward}}.
\newblock In \emph{{{ICLR19}}}, 2019.

\bibitem[Chaplot et~al.(2016)Chaplot, Sathyendra, Lample, and
  Salakhutdinov]{chaplot2016TransferDeepReinforcement}
Chaplot, D.~S., Sathyendra, K.~M., Lample, G., and Salakhutdinov, R.
\newblock Transfer {{Deep Reinforcement Learning}} in {{3D Environments}}: {{An
  Empirical Study}}.
\newblock In \emph{{{NIPS Deep Reinforcemente Leaning Workshop}}}, 2016.

\bibitem[Chen \& Mooney(2011)Chen and Mooney]{chenLearningInterpretNatural}
Chen, D.~L. and Mooney, R.~J.
\newblock Learning to {{Interpret Natural Language Navigation Instructions}}
  from {{Observation}}.
\newblock In \emph{{{AAAI11}}}, pp.\ ~7, 2011.

\bibitem[Chevalier-Boisvert(2018)]{gym_miniworld}
Chevalier-Boisvert, M.
\newblock gym-miniworld environment for openai gym.
\newblock \url{https://github.com/maximecb/gym-miniworld}, 2018.

\bibitem[Eysenbach et~al.(2019)Eysenbach, Gupta, Ibarz, and
  Levine]{eysenbach2019diayn}
Eysenbach, B., Gupta, A., Ibarz, J., and Levine, S.
\newblock Diversity is {{All You Need}}: {{Learning Skills}} without a {{Reward
  Function}}.
\newblock 2019.

\bibitem[Hausknecht \& Stone(2015)Hausknecht and Stone]{hausknecht2015rdqn}
Hausknecht, M. and Stone, P.
\newblock Deep {{Recurrent Q}}-{{Learning}} for {{Partially Observable MDPs}}.
\newblock 2015.

\bibitem[Hermann et~al.(2017)Hermann, Hill, Green, Wang, Faulkner, Soyer,
  Szepesvari, Czarnecki, Jaderberg, Teplyashin, Wainwright, Apps, Hassabis, and
  Blunsom]{hermann2017GroundedLanguageLearning}
Hermann, K.~M., Hill, F., Green, S., Wang, F., Faulkner, R., Soyer, H.,
  Szepesvari, D., Czarnecki, W.~M., Jaderberg, M., Teplyashin, D., Wainwright,
  M., Apps, C., Hassabis, D., and Blunsom, P.
\newblock Grounded {{Language Learning}} in a {{Simulated 3D World}}.
\newblock 2017.

\bibitem[Honnibal \& Montani(2017)Honnibal and Montani]{spacy2}
Honnibal, M. and Montani, I.
\newblock {spaCy 2}: Natural language understanding with {B}loom embeddings,
  convolutional neural networks and incremental parsing.
\newblock To appear, 2017.

\bibitem[Hutsebaut-Buysse et~al.(2020)Hutsebaut-Buysse, Mets, and
  Latré]{hutsebautbuysse2019fast}
Hutsebaut-Buysse, M., Mets, K., and Latré, S.
\newblock Fast {{Task-Adaptation}} for tasks labeled using {{Natural Language}}
  in {{Reinforcement Learning}}.
\newblock In \emph{{{ESANN2020}}}, 2020.

\bibitem[Jinnai et~al.(2020)Jinnai, Park, Machado, and
  Konidaris]{jinnai2020deep_covering_options}
Jinnai, Y., Park, J.~W., Machado, M.~C., and Konidaris, G.
\newblock Exploration in {{Reinforcement Learning}} with {{Deep Covering
  Options}}.
\newblock In \emph{{{ICLR2020}}}, pp.\ ~13, 2020.

\bibitem[Kapturowski et~al.(2019)Kapturowski, Ostrovski, Quan, Munos, and
  Dabney]{kapturowski2019r2d2}
Kapturowski, S., Ostrovski, G., Quan, J., Munos, R., and Dabney, W.
\newblock Recurrent {{Experience Replay}} in {{Distributed Reinforcement
  Learning}}.
\newblock 2019.

\bibitem[Lake et~al.(2017)Lake, Ullman, Tenenbaum, and
  Gershman]{lake2017building_ai}
Lake, B.~M., Ullman, T.~D., Tenenbaum, J.~B., and Gershman, S.~J.
\newblock Building machines that learn and think like people.
\newblock \emph{Behavioral and Brain Sciences}, 40, 2017.
\newblock ISSN 0140-525X, 1469-1825.
\newblock \doi{10.1017/S0140525X16001837}.

\bibitem[Luketina et~al.(2019)Luketina, Nardelli, Farquhar, Foerster, Andreas,
  Grefenstette, Whiteson, and Rocktäschel]{luketina2019suvery_rl_nlp}
Luketina, J., Nardelli, N., Farquhar, G., Foerster, J., Andreas, J.,
  Grefenstette, E., Whiteson, S., and Rocktäschel, T.
\newblock A {{Survey}} of {{Reinforcement Learning Informed}} by {{Natural
  Language}}.
\newblock In \emph{{{IJCAI19}}}, 2019.

\bibitem[Mei et~al.(2016)Mei, Bansal, and Walter]{mei2016ListenAttendWalk}
Mei, H., Bansal, M., and Walter, M.~R.
\newblock Listen, {{Attend}}, and {{Walk}}: {{Neural Mapping}} of
  {{Navigational Instructions}} to {{Action Sequences}}.
\newblock In \emph{{{AAAI16}}}, 2016.

\bibitem[Mikolov et~al.(2013)Mikolov, Chen, Corrado, and
  Dean]{mikolov2013word2vec}
Mikolov, T., Chen, K., Corrado, G., and Dean, J.
\newblock Efficient {{Estimation}} of {{Word Representations}} in {{Vector
  Space}}.
\newblock 2013.

\bibitem[Mnih et~al.(2015)Mnih, Kavukcuoglu, Silver, Rusu, Veness, Bellemare,
  Graves, Riedmiller, Fidjeland, Ostrovski, Petersen, Beattie, Sadik,
  Antonoglou, King, Kumaran, Wierstra, Legg, and Hassabis]{mnih2015dqn}
Mnih, V., Kavukcuoglu, K., Silver, D., Rusu, A.~A., Veness, J., Bellemare,
  M.~G., Graves, A., Riedmiller, M., Fidjeland, A.~K., Ostrovski, G., Petersen,
  S., Beattie, C., Sadik, A., Antonoglou, I., King, H., Kumaran, D., Wierstra,
  D., Legg, S., and Hassabis, D.
\newblock Human-level control through deep reinforcement learning.
\newblock 518\penalty0 (7540):\penalty0 529--533, 2015.
\newblock ISSN 0028-0836, 1476-4687.
\newblock \doi{10.1038/nature14236}.

\bibitem[Narasimhan et~al.(2018)Narasimhan, Barzilay, and
  Jaakkola]{narasimhan2018language_transfer}
Narasimhan, K., Barzilay, R., and Jaakkola, T.
\newblock Grounding {{Language}} for {{Transfer}} in {{Deep Reinforcement
  Learning}}.
\newblock 63, 2018.

\bibitem[Narvekar et~al.(2020)Narvekar, Peng, Leonetti, Sinapov, Taylor, and
  Stone]{narvekar2020curriculum_learning}
Narvekar, S., Peng, B., Leonetti, M., Sinapov, J., Taylor, M.~E., and Stone, P.
\newblock Curriculum {{Learning}} for {{Reinforcement Learning Domains}}: {{A
  Framework}} and {{Survey}}.
\newblock 2020.

\bibitem[Schrittwieser et~al.(2019)Schrittwieser, Antonoglou, Hubert, Simonyan,
  Sifre, Schmitt, Guez, Lockhart, Hassabis, Graepel, Lillicrap, and
  Silver]{schrittwieser2019muzero}
Schrittwieser, J., Antonoglou, I., Hubert, T., Simonyan, K., Sifre, L.,
  Schmitt, S., Guez, A., Lockhart, E., Hassabis, D., Graepel, T., Lillicrap,
  T., and Silver, D.
\newblock Mastering {{Atari}}, {{Go}}, {{Chess}} and {{Shogi}} by {{Planning}}
  with a {{Learned Model}}.
\newblock 2019.

\bibitem[Silver et~al.(2013)Silver, Yang, and Li]{silver2013lifelong_learning}
Silver, D.~L., Yang, Q., and Li, L.
\newblock Lifelong {{Machine Learning Systems}}: {{Beyond Learning
  Algorithms}}.
\newblock In \emph{{{AAAI13}}}, 2013.

\bibitem[Taylor \& Stone(2009)Taylor and Stone]{taylor2009transfer_rl_survey}
Taylor, M.~E. and Stone, P.
\newblock Transfer {{Learning}} for {{Reinforcement Learning Domains}}: {{A
  Survey}}.
\newblock 10, 2009.

\bibitem[Watkins(1989)]{watkins1989learning}
Watkins, C. J. C.~H.
\newblock Learning from delayed rewards.
\newblock 1989.

\bibitem[Weischedel et~al.(2013)Weischedel, Palmer, Marcus, Hovy, Pradhan,
  Ramshaw, Xue, Taylor, Kaufman, Franchini, El-Bachouti, Belvin, and
  Houston]{weischedelOntoNotesLargeTraining}
Weischedel, R., Palmer, M., Marcus, M., Hovy, E., Pradhan, S., Ramshaw, L.,
  Xue, N., Taylor, A., Kaufman, J., Franchini, M., El-Bachouti, M., Belvin, R.,
  and Houston, A.
\newblock {{OntoNotes}}: {{A Large Training Corpus}} for {{Enhanced
  Processing}}.
\newblock 2013.

\bibitem[Yosinski et~al.(2014)Yosinski, Clune, Bengio, and
  Lipson]{yosinski2014transfer_features}
Yosinski, J., Clune, J., Bengio, Y., and Lipson, H.
\newblock How transferable are features in deep neural networks?
\newblock In \emph{{{NIPS14}}}, 2014.

\bibitem[Zhong et~al.(2019)Zhong, Rocktäschel, and
  Grefenstette]{zhong2019RTFM}
Zhong, V., Rocktäschel, T., and Grefenstette, E.
\newblock {{RTFM}}: {{Generalising}} to {{Novel Environment Dynamics}} via
  {{Reading}}.
\newblock In \emph{{{ICLR20}}}, 2019.

\end{thebibliography}
\bibliographystyle{icml2020}

\end{document}